\documentclass[11pt]{article}

\usepackage[T1]{fontenc}
\usepackage[utf8]{inputenc}
\usepackage[polish,english]{babel}
\usepackage{times}
\usepackage{amssymb}

\usepackage[preprint]{acl}
\usepackage{amsmath}
\usepackage{latexsym}
\usepackage{amsthm}
\usepackage{bbm}
\usepackage{booktabs}
\usepackage{multirow}
\usepackage{graphicx}
\usepackage{rotating}
\usepackage{array}
\usepackage{cuted}
\usepackage{capt-of}
\usepackage{microtype}
\usepackage{inconsolata}

\usepackage[table,dvipsnames]{xcolor}
\usepackage[most]{tcolorbox}
\usepackage{arydshln}
\usepackage{standalone}

\usepackage{algorithm}
\usepackage{algpseudocode}
\usepackage{enumitem}

\setlist[itemize]{leftmargin=7pt, itemsep=-2pt, topsep=0pt}
\newtheorem{theorem}{Theorem}

\def\redc{\cellcolor[HTML]{FF999A}}
\def\orangec{\cellcolor[HTML]{FFCC99}}
\def\yellowc{\cellcolor[HTML]{FFF8AD}}

\newcommand{\ours}{PLR}

\title{\ours{}: \underline{P}lackett--\underline{L}uce for \underline{R}eordering In-Context Learning Examples}

\author{
\begin{tabular}{@{}c@{\hspace{2.5em}}c@{}}
Pawe{\l} Batorski & Paul Swoboda \\
\multicolumn{2}{c}{{\normalfont Heinrich Heine Universit\"at D\"usseldorf}} \\
\multicolumn{2}{c}{{\normalfont\texttt{\{pawel.batorski,paul.swoboda\}@hhu.de}}}
\end{tabular}
}

\begin{document}

\maketitle

\begin{strip}
\vspace{-2cm}

  \centering
  \includegraphics[width=\textwidth]{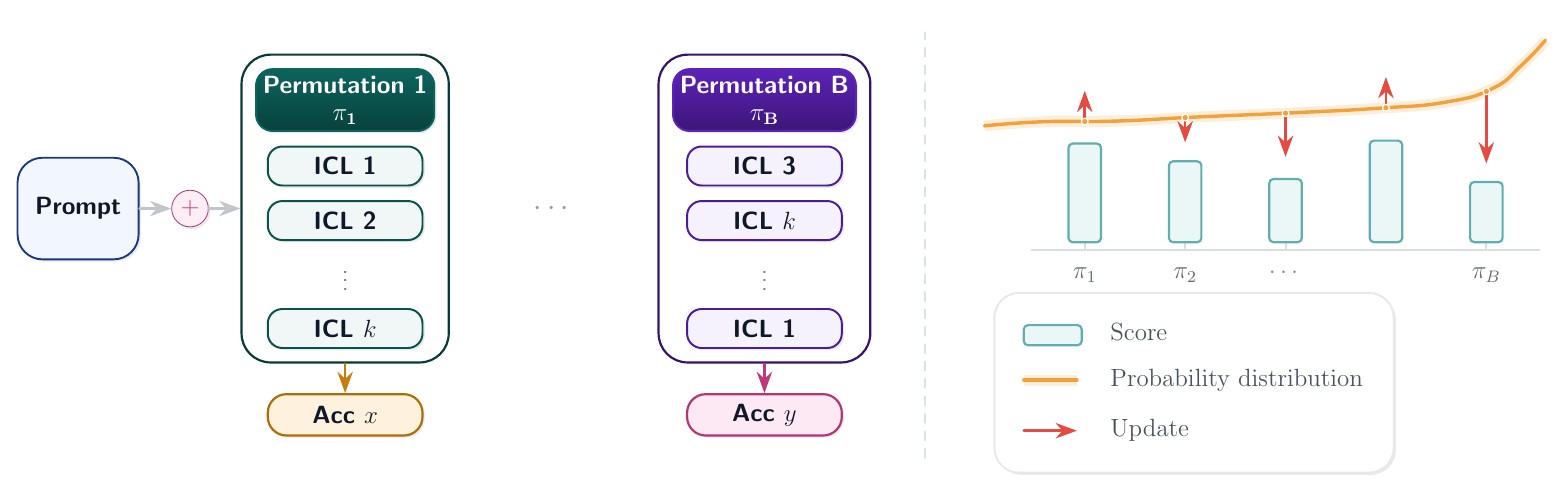}

\captionof{figure}{
Overview of our probabilistic approach to in-context example ordering. 
We maintain a Plackett-Luce distribution over permutations, repeatedly sample ICL example orderings, evaluate them under a task-level metric, and update the distribution to shift probability mass toward high-performing orders and away from low-performing ones.
}
 \label{fig:cycle_and_comparison}
\end{strip}

%\begin{strip}
%\vspace{-2cm}

%  \centering
%  \includegraphics[width=\textwidth]{teaser.pdf}

%\captionof{figure}{
%\swoboda{Tutaj nie jest wcale jasne, jaki problem my probujemy rozwiazac. Ilustracja musi byc bardziej abstrakcyjna. Tez te przyklady po prawej nie rozumiem.}
%Overview of \ours{}. We maintain a Plackett--Luce (PL) distribution over permutations of the in-context demonstrations and sample candidate orderings efficiently using the Gumbel perturb-and-sort (Gumbel--Top-$k$) trick. Each sampled permutation is evaluated on a split under a task-level metric, and the top-$K$ permutations are retained as an elite set. We then refit the PL parameters using one of three update rules: an exponential moving-average (EMA) rank update, maximum-likelihood estimation (MLE) on the elite permutations, or an EM-style update for mixture PL. The final output is an optimized distribution over orderings, from which we draw and select a single best permutation via Monte Carlo evaluation on a validation set. Unlike label-probability--based ordering methods, \ours{} directly optimizes the task metric and therefore applies to open-ended generation and reasoning tasks with unbounded answer spaces.
%}
% \label{fig:cycle_and_comparison}
%\end{strip}

%\begin{figure*}[t]
 % \centering
 % \vspace{-2cm} % adjust as needed
 % \includegraphics[width=\textwidth]{teaser.pdf}
 % \caption{Overview of \ours{}. ...}
 % \label{fig:cycle_and_comparison}
%\end{figure*}

\begin{abstract}
In-context learning (ICL) adapts large language models by conditioning on a small set of ICL examples, avoiding costly parameter updates.
Among other factors, performance is often highly sensitive to the \emph{ordering} of the examples.
However, exhaustive search over the $n!$ possible orderings is infeasible.
Therefore more efficient ordering methods use model confidence measures (e.g., label-probability entropy) over label sets or take a direct approach to finding the best ordering.
We propose \ours{}, a probabilistic approach to in-context example ordering that replaces discrete ordering search with learning a probability distribution over orderings with the Plackett--Luce model.
\ours{} models orderings using a Plackett--Luce distribution and iteratively updates its parameters to concentrate probability mass on high-performing orderings under a task-level metric.
Candidate orderings are sampled efficiently via a Gumbel perturb-and-sort procedure.
Experiments on multiple classification benchmarks show that \ours{} consistently improves few-shot accuracy for $k\!\in\!\{4,8,16,32\}$ examples, and we further demonstrate gains on mathematical reasoning tasks where label-based ordering methods are not applicable. Our code is available at: https://github.com/Batorskq/PLR. 
\end{abstract}

\section{Introduction}

Prompting LLMs  with in-context examples is a practical alternative to fine-tuning for adapting LLMs \citep{sahoo2024systematic, chen2023unleashing}.
While its mechanisms remain under active study, recent analyses suggest transformers can treat examples as a small inference-time ``training set,'' implicitly adapting from context \citep{akyurek2022learning, von2023transformers, xieexplanation}.
However, ICL can be brittle, minor semantics-preserving prompt edits often cause large performance changes \citep{razavi2025benchmarking, sclar2024quantifyinglanguagemodelssensitivity, chatterjee2024posixpromptsensitivityindex, zhuo2024prosaassessingunderstandingprompt, errica2024did, cao2024worstpromptperformancelarge}, including under natural perturbations and adversarially injected instructions \citep{zhu2024promptrobustevaluatingrobustnesslarge,  li2023evaluatinginstructionfollowingrobustnesslarge}. 
Another surprising aspect is that performance depends not only on which demonstrations are used but also on their order \citep{zhao2021calibrateuseimprovingfewshot, bhope2025optiseq, hehierarchical, lu2022fantasticallyorderedpromptsthem, min2022rethinkingroledemonstrationsmakes, reynolds2021prompt}.
Similar sensitivity appears in newer reasoning-oriented models \citep{guo2025deepseek}. 

The space of possible orders grows as $n!$, making exhaustive search for finding an optimal ordering infeasible.
Existing methods therefore rely on heuristics or limited candidate evaluation, including entropy- and label-distribution--based criteria such as LocalE/GlobalE~\citep{lu2022fantasticallyorderedpromptsthem} and PDO~\citep{xu2024context} and dataset-free filtering schemes like DEmO \citep{guo2024makes}.
These approaches work well for classification but typically require enumerating a finite label set (via verbalizers), which does not extend to open-ended generation and numerical reasoning with unbounded outputs (e.g., GSM8K). 
They also often perform brittle point-estimate in which a single ordering is sought, and may require factorial-time enumeration in small-$n$ regimes \citep{bhope2025optiseq}.
To address these limitations, we propose \ours{}, which replaces discrete permutation search with learning a distribution over orderings.
By modeling the space of orderings with a Plackett--Luce (PL) distribution (and mixtures thereof) and updating it to concentrate mass on high-performing regions, \ours{} robustly and efficiently explores the permutation space and directly optimizes the task metric without assuming a finite label space.

To summarize, our contributions are as follows:
\noindent
\begin{description}[leftmargin=!, labelwidth=10pt]
\item[Conceptual:]
We introduce \ours{}, a distributional approach to in-context example ordering that learns to place higher probability on better-performing example orderings under a task-level score.
Unlike prior label-probability heuristics that require enumerating a finite label set, \ours{} is label-space agnostic and applies naturally to open-ended generation and numerical reasoning.
We use the (mixture of) Plackett--Luce distribution and use Gumbel perturb-and-sort to sample efficiently from it.

\item[Algorithmic:]
We propose an iterative algorithm involving sampling for fitting the PL distribution, we propose three approaches: (i) a heuristic rank update, (ii) MLE and (iii) EM for the mixture of PL. We stabilize updates with EMA.

\item[Experimental:]
We evaluate \ours{} on classification and reasoning benchmarks with Qwen and Llama models, and provide ablations analyzing key design choices.
Our experiments show significantly higher performance than our baseline methods.
\end{description}

\section{Related Work}

\paragraph{Prompt Engineering and In-Context Prompt Optimization.}
Prompt engineering can substantially improve LLM performance without parameter updates~\citep{liu2023pretrain}, including reasoning-oriented methods such as Chain-of-Thought and self-consistency~\citep{wei2022chain,kojima2022large,wang2022self} and more structured variants (e.g., Tree-/Program-/Graph-of-Thought and decomposition)~\citep{yao2023tree,chen2022program,besta2024graph,zhou2023leasttomost}. Beyond manual design, automated prompt optimization searches over instructions and templates via model-guided refinement, evolutionary strategies, or black-box optimization~\citep{zhou2022large,pryzant2023automatic,guo12connecting,fernando2023promptbreeder,batorski2025gps,wang2023promptagent,yang2024llm_optimizers}, and related work optimizes discrete prompt tokens or uses strong random-search baselines~\citep{shin2020autoprompt,lu2024strings}. For in-context learning, performance also depends on which demonstrations are included and their \emph{ordering}~\citep{brown2020language,lu2022fantastically}, yet most prompt optimizers emphasize instruction text, treat demonstrations in a restricted way, or incur high cost when generating new examples~\citep{deng2022rlprompt,batorski2025prl, agarwal2024promptwizard, batorski2025piast, dziuba2026tatra}. We target this underexplored axis by optimizing the ordering of a fixed set of in-context examples.

\paragraph{ICL examples sensitivity.}
In-context learning can be fragile: even with the task and model fixed, small changes to a prompt with ICL examples e.g., which examples are used, their formatting/label strings, or their position in the context can cause large performance swings \citep{brown2020language, zhao2021calibrate, lu2024prompts, zhang2024batch}.
This sensitivity is often attributed to models relying on shallow patterns in examples (e.g., label/format heuristics) rather than task semantics \citep{min2022rethinking, fei2023mitigating, jang2024rectifying, sclar2023quantifying, chen2023relation, liangpearl, fangrethinking}.
Positional effects further amplify brittleness, with systematic biases toward certain prompt regions \citep{cobbina2025show, guo2025serial, xiang2024addressing} and reduced use of evidence in long contexts, especially for information in the middle \citep{liu2024lost}.
These findings motivate methods that explicitly optimize and robustify prompt construction, including ICL example ordering \citep{lu2024prompts, zhang2024batch}.

\paragraph{ICL Example Ordering Methods.}
A growing line of work studies how to order a fixed set of in-context examples to reduce brittleness. 
Prior analyses report strong positional and recency effects, where the same examples can perform very differently depending on their arrangement \citep{shin2022effect, min2022rethinking,wu2024prompt, rubin2022learning}.
Accordingly, many methods score or search over a subset of candidate orderings using proxy signals e.g., confidence/entropy heuristics, label-distribution matching, or other task-specific criteria  to avoid enumerating the full $n!$ space \citep{lu2022fantastically, xu2024context, guo2024makes, wang2023large, yang2023representative}.
More structured procedures, such as optimization-based search and rapid candidate refinement, can further improve reliability but still evaluate only a limited set of orders \citep{pham2025rapid, hehierarchical}.
In contrast, we perform distributional optimization over orderings to directly optimize the task metric.
This allows us to frame the optimization in a principled manner and, on a practical side, to explore efficiently the large underlying search space.

\section{Method} 
\label{sec:method}

\paragraph{Problem Statement.}
Let $E=\{(x_i,y_i)\}_{i=1}^n$ be a fixed set of $n$ in-context examples, where $x_i$ is an input/instruction and $y_i$ is its corresponding answer (label).
Let $S_n$ denote the symmetric group on $n$ elements, i.e., the set of all $n!$ orderings (or, mathematically, permutations) of $\{1,\dots,n\}$.
For any permutation $\pi=(\pi_1,\dots,\pi_n)\in S_n$, define the ordered example sequence
\[
E_\pi \;=\; \bigl((x_{\pi_1},y_{\pi_1}),\dots,(x_{\pi_n},y_{\pi_n})\bigr).
\]
Let $p$ be a fixed user-provided instruction (prefix prompt), and let $\oplus$ denote prompt concatenation.
The full prompt induced by $\pi$ is
\[
P(\pi) \;=\; p \oplus E_\pi .
\]
Given an evaluation dataset $D$ and a scoring function $f$ (e.g., accuracy) that evaluates the model on $D$ using prompt $P(\pi)$, our goal is to find
\[
\pi^\star \;=\; \arg\max_{\pi\in S_n}\; f\!\bigl(D,\; p \oplus E_\pi\bigr).
\]

\subsection{Model}
We will formulate finding the optimal ordering as fitting a distribution over permutations such that high quality permutations are given high probability mass.
To this end, we propose to use the Plackett--Luce (PL) distribution due to its computational efficiency and extend its expressive power by considering the mixture of PLs.

\paragraph{Framing example ordering as distribution estimation.}
Rather than searching directly for a single optimal ordering $\pi^\star\in S_n$, we maintain a parametric distribution $q_\phi(\pi)$ over permutations and optimize its parameters $\phi$ so that high-probability permutations achieve high task performance. Concretely, given an evaluation dataset $D$, a fixed prompt prefix $p$, and a scoring function $f(D, p \oplus E_\pi)$, we aim to maximize the expected score under $q_\phi$:
\begin{align}
\phi^\star
\;=\;
\arg\max_{\phi}\;
\mathbb{E}_{\pi \sim q_\phi}\Bigl[ f\!\bigl(D,\; p \oplus E_\pi\bigr) \Bigr].
\label{eq:dist_objective}
\end{align}
In our setting, $q_\phi$ is instantiated as a Plackett--Luce distribution, which provides an efficient mechanism to sample candidate permutations and progressively concentrate probability mass on better-performing orderings. 

\paragraph{Plackett--Luce distribution.}
The Plackett--Luce (PL) model \citep{luce1959individual, plackett1975analysis} defines a probability distribution over permutations $\pi=(\pi_1,\dots,\pi_n)\in S_n$ parameterized by a score (logit) vector $\theta\in\mathbbmss{R}^n$.
The model can be interpreted as a sequential choice process: at rank position $r$, the item placed at that position is drawn from the set of remaining items with probability proportional to its exponentiated score.
Formally,
\begin{align}
\Pr(\pi \mid \theta)
&=
\prod_{r=1}^{n}
\frac{\exp\!\left(\theta_{\pi_r}\right)}
{\sum\limits_{j\in R_r}\exp\!\left(\theta_j\right)},
\label{eq:pl}
\end{align}
where $R_r = \{\pi_r,\pi_{r+1},\dots,\pi_n\}$ denotes the set of items not yet selected at step $r$. The PL distribution is invariant to additive shifts of the logits, i.e., $\Pr(\pi\mid\theta)=\Pr(\pi\mid\theta+c\mathbf{1})$ for any $c\in\mathbbmss{R}$, and it is therefore common to impose an identifiability constraint such as $\sum_{i=1}^n \theta_i = 0$.

\paragraph{Mixture of PL distributions.}
A single PL model uses one global logit vector $\theta$ to generate a ranking via sequential choices.
Although the selection is conditioned on previously chosen items, this conditioning operates only through removal: the same logits are reused at every step on the remaining set.
In particular, for any two items $a,b$ that are still available at position $r$, PL satisfies an independence-of-irrelevant-alternatives (IIA) property~\citep{luce1959individual,plackett1975analysis},
\[
\frac{\Pr(\pi_r=a \mid \pi_{<r};\theta)}{\Pr(\pi_r=b \mid \pi_{<r};\theta)}
=
\exp(\theta_a-\theta_b),
\]
which is invariant to the previously selected items $\pi_{<r}$ (except that selected items are excluded). Consequently, a single PL component cannot express interaction effects such as ``if example $i$ is placed first, then example $j$ should be placed second'' beyond the trivial fact that $i$ is no longer eligible. To partially relax this limitation while retaining efficient sampling, we consider a mixture of PL distributions \citep{zhao2016learning, zhao2019learning}:
$$
q(\pi) \;=\; \sum_{k=1}^{K} \alpha_k\, \mathrm{PL}\!\left(\pi \mid \theta^{(k)}\right),
$$
where $\alpha_k\ge 0,\;\; \sum_{k=1}^{K}\alpha_k=1.$
The mixture introduces a latent variable $z\in\{1,\dots,K\}$ that indicates from which mixture component the permutation was drawn. 
The mixture of PL distribution allows to model distributions that are not subject to the IIA condition, even though each individual mixture distribution does so.
This increases expressivity for multi-modal sets of high-performing demonstration orders while preserving the same perturb-and-sort sampling mechanism within each component.

%Importantly, early selections update the posterior over components, which in turn changes what is likely to be placed next.

%For instance, after selecting $\pi_1=i$,
%\begin{align}
%\mathrm{P}(\pi_2=j \mid \pi_1=i)
%&=
%\sum_{k=1}^{K} \mathrm{P}(z=k \mid \pi_1=i)\, \nonumber\\
%&\qquad\cdot
%\frac{\exp(\theta^{(k)}_j)}{\sum_{u\neq i}\exp(\theta^{(k)}_u)}.
%\label{eq:mixpl_second_step}
%\end{align}
%where $\Pr(z=k \mid \pi_1=i)\propto \alpha_k\,\Pr(\pi_1=i\mid \theta^{(k)})$, thus breaking the IIA condition.
%Therefore, even though each PL component individually satisfies IIA, the mixture can represent context-dependent preferences (e.g., ``if $i$ is first, then $j$ is likely second'') by shifting probability mass toward components whose ordering patterns are consistent with the early choices.

Even stronger, it is known that mixture of PL can approximate any distribution over permutations arbitrarily well when the number of mixture components $K$ is large enough, see Theorem~\ref{thm:expressivity-mixture-PL} in the appendix.

\paragraph{Gumbel trick for sampling PL permutations.}

Sampling from a Plackett--Luce model can be done via sequential softmax choices over the remaining items, but this procedure is inherently sequential and can be inefficient when drawing many permutations, therefore we use the Gumbel Trick \citep{kool2019stochastic,maddison2014sampling} which simplifies the sampling problem to sampling from a Gumbel distribution.
Let $\theta\in\mathbb{R}^n$ be the PL logits, and draw i.i.d.\ Gumbel noise variables
\[
g_i \sim \mathrm{Gumbel}(0,1), \qquad i=1,\dots,n,
\]
which can be obtained from $u_i\sim \mathrm{Uniform}(0,1)$ by
\[
g_i = -\log\!\bigl(-\log(u_i)\bigr).
\]
Define perturbed scores $s_i = \theta_i + g_i$ and sort items by decreasing $s_i$. Denoting by $\pi$ the resulting ordering (i.e., $s_{\pi_1}\ge s_{\pi_2}\ge\cdots\ge s_{\pi_n}$), the induced random permutation satisfies
\[
\pi \sim \mathrm{PL}(\theta).
\]

Intuitively, the Gumbel perturbations transform sampling from a sequence of softmax choices into a single ``perturb-and-sort'' operation, enabling efficient and numerically stable generation of full rankings from the PL model.

For the mixture of PL we first sample the mixture component using the mixture weights $\alpha_i$, after which we sample as above from the selected PL distribution.

\begin{figure}
    \centering
    \includegraphics[width=\columnwidth]{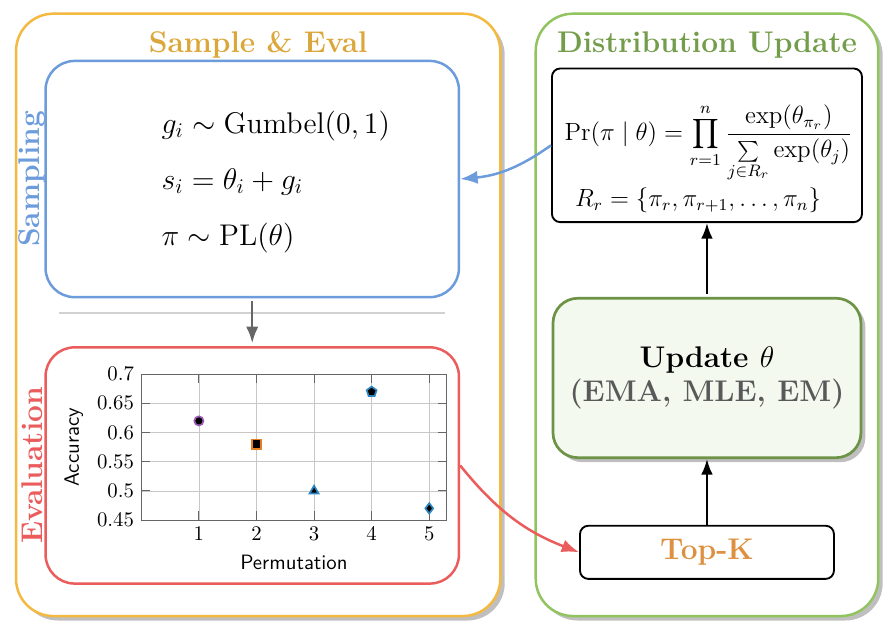}
    \caption{Illustration of \ours{}. Given a Plackett--Luce distribution, we sample high-probability permutations with the Gumbel trick. Each permutation is scored and the top-K are retained and used for fitting an improved Plackett-Luce distribution. This is iterated until high quality ICL ordering is found.}
    \label{fig:alg}
\end{figure}

\subsection{Training}
We illustrate our approach to finding an ordering via fitting a PL distribution in Algorithm~\ref{alg:pseudo-code}.
We start with a uniform initial distribution (line~\ref{alg:init}), sample a number of permutations (lines~\ref{alg:sample-start}-\ref{alg:sample-end}) and evaluate their performance on the training set (line~\ref{alg:eval-training}), fit the distribution to the best permutations according to the training set (line~\ref{alg:fit}) and iterate. 
The final permutation is obtained by sampling several permutations from the obtained distribution (line~\ref{alg:sample-val}) and taking the best according to a held-out validation set (line~\ref{alg:best-val}).
The extension to mixture of PL is given straightforwardly by replacing all PL specific steps by the mixture of PL counterparts.
A graphical illustration of the overall procedure is given in Figure~\ref{fig:alg}.

We propose three ways to fit the (mixture of) PL distributions below. 

\begin{algorithm}[ht]
\caption{\ours{}}
\label{alg:pl_generic}
\small
\begin{algorithmic}[1]
\Require ICL examples $E=\{(x_i,y_i)\}_{i=1}^n$, prefix prompt $p$, metric $f(\cdot)$, model $\mathcal{M}$
\Require training sets $\{D^{\mathrm{train}}_t\}_{t=1}^{T}$, validation set $D^{\mathrm{val}}$
\Require CE params: batch $B$, elite fraction $\rho$, final draws $K$
\Require update routine $\textsc{Update}(\theta;\mathcal{E}_t,\{s^{(b)}\}_{\pi^{(b)}\in\mathcal{E}_t},\ldots)$
\State $\theta \gets \mathbf{0}\in\mathbb{R}^n$ \Comment{PL logits: enforce $\sum_i\theta_i=0$}
\For{$t=1,\dots,T$}
\label{alg:init}

  \Statex \textbf{Sample $B$ permutations via Gumbel perturb-and-sort:}
  \For{$b=1,\dots,B$}
    \For{$i=1,\dots,n$}
\label{alg:sample-start}  
      \State Sample $u_i \sim \mathrm{Uniform}(0,1)$
      \State $g_i \gets -\log\!\bigl(-\log(u_i)\bigr)$ \Comment{$g_i \sim \mathrm{Gumbel}(0,1)$}
      \State $\tilde{s}_i \gets \theta_i + g_i$
    \EndFor
    \State $\pi^{(b)} \gets \mathrm{argsort}_{i\in\{1,\dots,n\}}\bigl(\tilde{s}_i\bigr)$ (descending)
  \EndFor
\label{alg:sample-end} 

  \State Score $s^{(b)} \gets f\!\bigl(D^{\mathrm{train}}_t,\; p \oplus E_{\pi^{(b)}};\; \mathcal{M}\bigr)$ for all $b$
  \label{alg:eval-training}
  \State Let $\mathcal{E}_t \gets \textsc{Top}_{\lceil \rho B\rceil}\bigl(\{\pi^{(b)}\}, \{s^{(b)}\}\bigr)$
  \label{alg:elite-set-selection}
  \State $\theta \gets \textsc{Update}\!\left(\theta;\; \mathcal{E}_t,\; \{s^{(b)}\}_{\pi^{(b)}\in\mathcal{E}_t},\ldots\right)$
  \label{alg:fit}
  \State $\theta \gets \theta - \frac{1}{n}\sum_{i=1}^n \theta_i$ \Comment{center logits}

\EndFor
\State Draw $\{\pi^{(k)}\}_{k=1}^{K}\sim \mathrm{PL}(\cdot\mid\theta)$ (same Gumbel trick) and select
\label{alg:sample-val}
\State $\hat{\pi} \gets \arg\max_{\pi^{(k)}} f\!\bigl(D^{\mathrm{val}},\; p \oplus E_{\pi^{(k)}};\; \mathcal{M}\bigr)$
\State \Return $\hat{\pi}$
\label{alg:best-val}
\end{algorithmic}
\label{alg:pseudo-code}
\end{algorithm}

\paragraph{Training the Plackett--Luce parameters.}
%Our optimization follows an \swoboda{estimation-of-distribution / cross-entropy style} procedure \swoboda{Jako czesc odnosi sie do estimate of distribution, jako do cross entropy w tekscie ponizej?}: we iteratively sample permutations, keep an elite set under the task score, and refit the sampling distribution to increase mass on high-performing regions \citep{rubinstein1999cross,de2005tutorial}.
We propose three ways to estimate PL parameters: a heuristic rank update and using MLE for the PL distribution and EM for the mixture of PL.
We stabilize all by using exponential moving averages.

The learning happens on line~\ref{alg:fit} of Algorithm~\ref{alg:pseudo-code}.
We are given the current PL logits $\theta\in\mathbb{R}^n$ (and, for mixtures, weights) and  a set of high-scoring (\emph{elite}) permutations produced during the optimization procedure $\mathcal{E}=\{\pi^{(1)},\dots,\pi^{(M)}\}\subset S_n$, optionally with nonnegative weights $\{w_m\}_{m=1}^M$ (e.g., proportional to training accuracy).

\smallskip
\noindent\textbf{(i) Exponential moving average (EMA) / heuristic rank update.}
A simple update constructs a target score vector from the average elite ranks. Let $\mathrm{rank}_{\pi}(i)\in\{0,\dots,n-1\}$ be the position of item $i$ in permutation $\pi$ (higher logits correspond to earlier position in the permutation). Define the elite average rank
\[
\bar{r}_i \;=\; \frac{1}{M}\sum_{m=1}^M \mathrm{rank}_{\pi^{(m)}}(i),
\]
and a rank-to-logit target (with temperature $\tau>0$)
\[
\theta^{\mathrm{tgt}}_i \;=\; -\frac{\bar{r}_i}{\tau}.
\]
Thus, if the average elite rank is low (i.e.\ example occurs early), then its logits should be increased in comparison to other logits (corresponding to small $\bar{r}_i/\tau$). Vice versa for high rank.

We then apply an EMA-style update with step size $\alpha\in[0,1]$:
\[
\theta \;\leftarrow\; (1-\alpha)\,\theta \;+\; \alpha\,\theta^{\mathrm{tgt}}.
\]
Finally, we enforce identifiability and numerical stability by centering and clipping, e.g.\ $\theta \leftarrow \theta - \frac{1}{n}\sum_{i=1}^n\theta_i$ and $\theta_i\in[-c,c]$.

\smallskip
\noindent\textbf{(ii) Maximum likelihood (MLE) on elite permutations.}
A more principled alternative fits $\theta$ by maximizing the (weighted) PL log-likelihood of the elite permutations:
\[
\theta^{\mathrm{MLE}}
\;\in\;
\arg\max_{\theta\in\mathbb{R}^n}
\sum_{m=1}^M w_m \log \Pr(\pi^{(m)}\mid \theta),
\]
where $\Pr(\pi\mid\theta)$ is given in Eq.~\eqref{eq:pl}. In practice we solve this optimization with gradient-based methods (e.g.\ Adam), again applying a centering constraint to resolve the additive invariance of $\theta$. To stabilize optimization across iterations, we optionally combine MLE with an EMA step:
\[
\theta \;\leftarrow\; (1-\alpha)\,\theta \;+\; \alpha\,\theta^{\mathrm{MLE}}.
\]
%This yields a CE-style update in which the next sampling distribution is the PL model that best matches the elite set under maximum likelihood.
The MLE is well-behaved due to the Plackett--Luce log-likelihood being concave in $\theta$ \citep{maystre2015fast}.

% \paragraph{Selecting the final permutation.}
% After optimizing the permutation distribution $q_{\phi^\star}(\pi)$, we obtain a single ordering by Monte Carlo selection. Specifically, we draw \swoboda{$K$ candidate permutations <- ilosc komponentow mieszankowych}
% \[
% \pi^{(1)},\dots,\pi^{(K)} \;\sim\; q_{\phi^\star}(\pi),
% \]
% evaluate each candidate on the validation set $D_{\mathrm{val}}$ using the scoring function $f$, and select the best-performing permutation:
% \[
% \hat{\pi}
% \;=\;
% \arg\max_{k\in\{1,\dots,K\}}
% f\!\bigl(D_{\mathrm{val}},\; p \oplus E_{\pi^{(k)}}\bigr).
% \]

\begin{table*}[t]
\centering
\scriptsize
\setlength{\tabcolsep}{3pt}
\renewcommand{\arraystretch}{1.49}

\caption{Test accuracy (\%) averaged over seeds. best score is colored in red, second best in orange and third best in yellow. Avg is macro-average over MR/NEWS/SST5/TREC/SUBJ. Top block: \texttt{qwen2.5-7b-instruct}. Bottom block: \texttt{Llama3.1-8B-instruct}.}
\label{tab:cls_results_k_qwen_llama_sibling}

\resizebox{\textwidth}{!}{%
\begin{tabular}{l c *{24}{c}}
\toprule
\multirow{2}{*}{Method} & \multirow{2}{*}{Model}
& \multicolumn{4}{c}{MR}
& \multicolumn{4}{c}{NEWS}
& \multicolumn{4}{c}{SST5}
& \multicolumn{4}{c}{TREC}
& \multicolumn{4}{c}{SUBJ}
& \multicolumn{4}{c}{Avg} \\
\cmidrule(lr){3-6}\cmidrule(lr){7-10}\cmidrule(lr){11-14}\cmidrule(lr){15-18}\cmidrule(lr){19-22}\cmidrule(lr){23-26}
&& $k{=}4$ & $k{=}8$ & $k{=}16$ & $k{=}32$
& $k{=}4$ & $k{=}8$ & $k{=}16$ & $k{=}32$
& $k{=}4$ & $k{=}8$ & $k{=}16$ & $k{=}32$
& $k{=}4$ & $k{=}8$ & $k{=}16$ & $k{=}32$
& $k{=}4$ & $k{=}8$ & $k{=}16$ & $k{=}32$
& $k{=}4$ & $k{=}8$ & $k{=}16$ & $k{=}32$ \\
\midrule
Static
& \multirow{11}{*}{\rotatebox{90}{\texttt{Qwen2.5-7B-Instruct}}}
& 90.22 & 91.10 & 91.75 & 91.73 & 83.37 & 82.47 & 83.71 & 83.79 & 54.78 & 54.12 & 54.78 & 56.52 & 61.95 & 63.90 & 60.26 & 59.99 & 65.88 & 71.50 & 76.55 & 79.23 & 71.24 & 72.62 & 73.41 & 74.25 \\

\texttt{Top-K}
&
& \redc 90.90 & 91.14 & \redc 91.95 & 91.96 & 84.18 & 84.40 & 85.07 & 85.35 & \orangec 55.55 & 55.36 & 55.85 & 56.85 & \yellowc 65.79 & 65.45 & 61.00 & \orangec 63.29 & 71.54 & 77.43 & 79.69 & 82.32 & 73.59 & 74.76 & 74.71 & 75.95 \\

\texttt{LocalE}
&
& 90.72 & 91.20 & 91.36 & 91.90 & 82.59 & 83.53 & 83.69 & 82.76 & 54.72 & 54.55 & 54.67 & 56.86 & 63.87 & 65.86 & 61.14 & 61.74 & 69.07 & 73.70 & 75.96 & 80.05 & 72.19 & 73.77 & 73.36 & 74.66 \\

\texttt{GlobalE}
&
& 90.66 & 91.11 & 91.69 & 91.80 & 82.79 & 83.60 & 83.81 & 82.97 & 54.50 & 54.47 & 55.07 & 56.53 & 59.31 & 63.02 & 61.67 & 61.07 & 68.48 & 70.93 & 74.64 & 77.82 & 71.15 & 72.63 & 73.38 & 74.04 \\

\texttt{PDO}
&
& 90.56 & 91.08 & 91.39 & 91.73 & 84.11 & 83.50 & 82.64 & 84.23 & 54.87 & 54.34 & 55.31 & 56.29 & 59.01 & 64.10 & 62.21 & 61.81 & 65.69 & 70.12 & 75.26 & 77.00 & 70.85 & 72.63 & 73.36 & 74.21 \\

\texttt{PDO-U}
&
& 90.61 & 90.91 & \orangec 91.87 & 91.85 & 83.94 & 84.06 & 84.73 & 85.13 & 54.64 & 55.51 & 55.28 & \orangec 57.32 & 60.33 & 64.03 & \orangec 63.49 & \yellowc 63.16 & \orangec 72.83 & 76.79 & 78.82 & 81.23 & 72.47 & 74.26 & 74.84 & 75.74 \\

\texttt{PDO-Up}
&
& 90.61 & 91.10 & \orangec 91.87 & 91.83 & 83.79 & 83.98 & 84.73 & 85.34 & 54.76 & \orangec 55.60 & 55.76 & \yellowc 57.28 & 59.32 & 65.32 & \yellowc 63.22 & 62.08 & \orangec 72.83 & 76.79 & 78.82 & 81.23 & 72.26 & 74.56 & 74.88 & 75.55 \\

\texttt{DeMO}
&
& \yellowc 90.79 & 91.17 & 91.67 & 91.70 & 83.50 & 83.91 & 83.66 & 84.77 & 55.37 & 54.47 & 55.66 & 57.06 & 63.06 & \yellowc 66.13 & \redc 63.97 & \redc 65.59 & 68.95 & 72.06 & 74.15 & 78.94 & 72.33 & 73.55 & 73.82 & 75.61 \\

\cmidrule(lr){1-1}
\cmidrule(lr){3-26}

\texttt{\ours{}-EMA}
&
& 90.67 & \redc 91.46 & \yellowc 91.83 & \redc 92.21 & \orangec 84.30 & \yellowc 84.80 & \orangec 86.14 & \orangec 86.38 & \yellowc 55.38 & 55.41 & \yellowc 56.74 & 57.18 & 65.29 & \orangec 66.60 & 62.08 & 62.15 & 72.78 & \orangec 80.64 & \orangec 83.98 & \redc 86.75 & \yellowc 73.68 & \orangec 75.78 & \orangec 76.15 & \redc 76.93 \\

\texttt{\ours{}-1}
&
& 90.78 & \yellowc 91.32 & \redc 91.95 & \orangec 92.14 & \redc 84.34 & \orangec 84.89 & \yellowc 85.80 & \redc 86.82 & 55.08 & \yellowc 55.55 & \orangec 56.78 & 56.86 & \redc 66.91 & 65.99 & 61.47 & 61.20 & \yellowc 72.80 & \yellowc 80.33 & \yellowc 83.44 & \yellowc 85.90 & \orangec 73.98 & \yellowc 75.62 & \yellowc 75.89 & \yellowc 76.58 \\

\texttt{\ours{}-4}
&
& \orangec 90.82 & \orangec 91.40 & 91.79 & \yellowc 92.02 & \yellowc 84.29 & \redc 84.97 & \redc 86.28 & \yellowc 85.98 & \redc 55.58 & \redc 56.04 & \redc 57.18 & \redc 57.81 & \orangec 66.60 & \redc 66.94 & 62.28 & 62.08 & \redc 73.09 & \redc 80.94 & \redc 85.17 & \orangec 85.98 & \redc 74.08 & \redc 76.06 & \redc 76.54 & \orangec 76.77 \\

\midrule\midrule

\texttt{Static}
& \multirow{11}{*}{\rotatebox{90}{\texttt{Llama3.1-8B-Instruct}}}
& 91.34 & 91.87 & 92.07 & 92.39 & 79.18 & 80.99 & 82.59 & 83.02 & 50.14 & 49.50 & 51.93 & 53.57 & 53.12 & 58.95 & 56.80 & 65.28 & 79.76 & 84.35 & 84.12 & 87.17 & 70.71 & 73.13 & 73.50 & 76.29 \\

\texttt{Top-K}
&
& 91.49 & 91.99 & 92.44 & 93.11 & \yellowc 81.82 & 83.66 & 85.00 & 85.30 & \yellowc 51.27 & 53.91 & 54.84 & 54.33 & 56.38 & \yellowc 63.35 & 62.34 & 66.25 & 85.61 & 86.44 & 90.12 & 91.15 & 73.31 & 75.87 & 76.95 & 78.03 \\

\texttt{LocalE}
&
& 91.53 & 91.95 & 91.87 & 92.71 & 80.11 & 81.35 & 83.27 & 82.72 & 50.48 & 51.17 & 52.50 & 54.08 & \redc 58.01 & 60.96 & 58.01 & 58.15 & 81.41 & 81.04 & 82.37 & 87.54 & 72.31 & 73.29 & 73.60 & 75.04 \\

\texttt{GlobalE}
&
& 91.42 & 91.90 & 91.92 & 92.75 & 78.64 & 81.62 & 82.70 & 84.62 & 50.14 & 52.30 & 53.69 & 53.67 & 55.78 & 60.61 & 61.24 & 65.36 & 82.56 & 83.19 & 83.71 & 89.48 & 71.71 & 73.92 & 74.65 & 77.18 \\

\texttt{PDO}
&
& 91.09 & 91.84 & 92.22 & 92.81 & 80.47 & 82.42 & 84.65 & 84.42 & 50.42 & 52.64 & 54.40 & 54.35 & 56.07 & 60.05 & 56.42 & 64.98 & 83.11 & 83.84 & 87.26 & 89.97 & 72.23 & 74.16 & 74.99 & 77.31 \\

\texttt{PDO-U}
&
& 91.36 & 92.26 & 92.48 & 92.86 & 81.75 & 83.37 & 84.22 & 84.92 & 49.62 & 52.43 & 53.94 & 53.13 & 54.11 & 61.22 & 59.94 & 60.14 & 85.52 & 86.12 & 88.58 & 90.55 & 72.47 & 75.08 & 75.83 & 76.32 \\

\texttt{PDO-Up}
&
& 91.43 & 92.26 & 92.46 & 92.96 & 81.75 & 83.37 & 84.48 & 84.92 & \orangec 51.60 & 54.35 & 54.46 & 54.49 & 55.63 & 62.14 & 61.98 & 68.25 & 85.56 & 86.12 & 88.01 & 90.63 & 73.19 & 75.65 & 76.28 & 78.25 \\

\texttt{DeMO}
&
& \orangec 91.62 & \orangec 92.33 & 92.40 & 93.05 & 80.88 & 82.49 & 84.59 & 84.34 & \redc 51.64 & 52.84 & 53.60 & 54.22 & 54.98 & 61.06 & 62.25 & 67.05 & 78.15 & 83.37 & 87.89 & 90.25 & 71.45 & 74.42 & 76.15 & 77.78 \\ 

\cmidrule(lr){1-1}
\cmidrule(lr){3-26}

\texttt{\ours{}-EMA}
&
& 91.42 & \yellowc 92.28 & \yellowc 92.69 & \redc 93.24 & \orangec 81.93 & \yellowc 84.47 & \orangec 86.20 & \yellowc 85.96 & 51.09 & \redc 55.44 & \yellowc 55.08 & \yellowc 55.04 & \orangec 57.31 & \redc 63.81 & \yellowc 62.94 & \redc 71.54 & \yellowc 86.40 & \yellowc 89.35 & \yellowc 92.26 & \yellowc 93.40 & \orangec 73.63 & \orangec 77.07 & \yellowc 77.83 & \orangec 79.84 \\

\texttt{\ours{}-1}
&
& \redc 91.70 & \redc 92.39 & \orangec 92.79 & \orangec 93.19 & \redc 82.13 & \redc 85.23 & \redc 86.37 & \orangec 86.19 & 51.04 & \yellowc 54.83 & \orangec 55.17 & \orangec 55.64 & \yellowc 57.26 & \orangec 63.80 & \orangec 64.99 & \yellowc 69.93 & \redc 86.46 & \redc 90.59 & \orangec 92.45 & \redc 94.19 & \redc 73.72 & \redc 77.37 & \orangec 78.35 & \yellowc 79.83 \\

\texttt{\ours{}-4}
&
& \yellowc 91.54 & 92.16 & \redc 92.80 & \yellowc 93.13 & 81.52 & \orangec 84.69 & \yellowc 86.10 & \redc 86.31 & 51.03 & \orangec 54.99 & \redc 55.22 & \redc 55.96 & 57.14 & 62.82 & \redc 65.97 & \orangec 70.63 & \orangec 86.44 & \orangec 89.69 & \redc 93.22 & \orangec 93.59 & \yellowc 73.53 & \yellowc 76.87 & \redc 78.66 & \redc 79.92 \\

\bottomrule
\end{tabular}%
}
\end{table*}

\paragraph{EM-style training for the mixture.}
We fit the mixture parameters $(\alpha,\{\theta^{(k)}\}_{k=1}^K)$ using an EM-style procedure on the current elite set of permutations $\mathcal{E}=\{\pi^{(m)}\}_{m=1}^M$, optionally weighted by $w_m$ (e.g., proportional to training accuracy). In the E-step, we compute responsibilities
\[
r_{mk}
=
\frac{\alpha_k\,\mathrm{PL}(\pi^{(m)}\mid \theta^{(k)})}{\sum_{\ell=1}^{K}\alpha_\ell\,\mathrm{PL}(\pi^{(m)}\mid \theta^{(\ell)})}.
\]
In the M-step, we update mixture weights via $\alpha_k \leftarrow \frac{\sum_m w_m r_{mk}}{\sum_m w_m}$ and update each component logits by weighted maximum likelihood,
\[
\theta^{(k)}
\leftarrow
\arg\max_{\theta}\sum_{m=1}^{M} w_m r_{mk}\log \mathrm{PL}(\pi^{(m)}\mid \theta),
\]
which we optimize approximately with a few steps of Adam and then optionally smooth with an EMA update for stability. This EM-style update allows the mixture to allocate different components to different modes of high-performing orderings while preserving efficient PL sampling within each component.

\section{Experiments}
We follow the standard evaluation protocol in prior work on ICL example ordering.
For each dataset and each number of ICL examples $k \in \{4,8,16,32\}$ we run 5 random seeds.
In every seed, we (i) sample a fresh set of $k$ demonstrations to include in the prompt,
(ii) sample disjoint training and validation sets used only for training/selecting an ordering, 
and (iii) report performance on the (held-out) test set using the selected ordering.
Additionally, we focus on the ordering of examples rather than their selection, ensuring that all methods use the same ICL examples so that the effects of ordering can be isolated. This strategy prevents unfair comparisons that could result from using different sets of examples.
Our ablations are conducted on the SUBJ dataset, which is particularly sensitive to the choice and ordering of ICL examples.

\begin{table*}[ht]
\centering
\scriptsize
\setlength{\tabcolsep}{3pt}
\renewcommand{\arraystretch}{1.5}

\caption{Test accuracy (\%) averaged over seeds for \texttt{Qwen2.5-7B-instruct} on math benchmarks. \redc{} best, \orangec{} second best, \yellowc{} third best (per column). Avg is macro-average over GSM8K/DeepMath/Math500.}
\label{tab:math_results_k_qwen}
\resizebox{\textwidth}{!}{%
\begin{tabular}{l *{16}{c}}
\toprule
\multirow{2}{*}{Method}
& \multicolumn{4}{c}{GSM8K}
& \multicolumn{4}{c}{DeepMath}
& \multicolumn{4}{c}{Math500}
& \multicolumn{4}{c}{Avg} \\
\cmidrule(lr){2-5}\cmidrule(lr){6-9}\cmidrule(lr){10-13}\cmidrule(lr){14-17}
& $k{=}4$ & $k{=}8$ & $k{=}16$ & $k{=}32$
& $k{=}4$ & $k{=}8$ & $k{=}16$ & $k{=}32$
& $k{=}8$ & $k{=}4$ & $k{=}32$ & $k{=}16$
& $k{=}4$ & $k{=}8$ & $k{=}16$ & $k{=}32$ \\
\midrule

\texttt{Static}
& 35.45 & 36.33 & \yellowc 39.70 & 39.05
& \yellowc 36.03 & 38.70 & 41.85 & 43.27
& 28.52 & 30.83 & 29.60 & 34.98
& 34.10 & 34.52 & 38.84 & 37.31 \\

\texttt{Top-K}
& \yellowc 36.05 & 38.86 & 39.35 & 40.40
& \orangec 36.49 & 38.74 & 42.20 & 45.13
& 31.40 & \yellowc 31.13 & 29.80 & \yellowc 38.60
& \yellowc 34.56 & 36.33 & 40.05 & 38.44 \\

\hline

\texttt{\ours{}-EMA}
& \redc 40.88 & \yellowc 39.50 & \orangec 39.85 & \yellowc 41.60
& \redc 36.64 & \redc 41.52 & \orangec 42.52 & \orangec 46.16
& \redc 32.32 & \redc 33.73 & \yellowc 31.80 & \redc 39.00
& \redc 37.08 & \redc 37.78 & \yellowc 40.46 & \yellowc 39.85 \\

\texttt{\ours{}-1}
& \orangec 39.63 & \orangec 39.65 & \orangec 39.85 & \redc 42.85
& \redc 36.64 & \orangec 39.93 & \redc 42.69 & \yellowc 45.62
& \yellowc 31.43 & \orangec 32.70 & \orangec 32.00 & \redc 39.00
& \orangec 36.32 & \orangec 37.00 & \orangec 40.51 & \orangec 40.16 \\

\texttt{\ours{}-4}
& \redc 40.88 & \redc 39.77 & \redc 41.40 & \orangec 42.80
& \redc 36.64 & \yellowc 39.17 & \yellowc 42.46 & \redc 46.36
& \orangec 31.90 & \redc 33.73 & \redc 33.40 & \orangec 38.78
& \redc 37.08 & \yellowc 36.95 & \redc 40.88 & \redc 40.85 \\

\bottomrule
\end{tabular}%
}
\end{table*}

\subsection{Baselines}
In this section we summarize the baselines we compare \texttt{\ours{}} against.

\begin{itemize}
    \item \texttt{Static.}
    We keep the $k$ sampled demonstrations in their original (data) order, i.e., no reordering.

    \item \texttt{Top-K}.
    We uniformly sample candidate permutations and evaluate them on the validation set.
    We use the same number of sampled candidates as \texttt{\ours{}}.
    
%    \swoboda{Czy tutaj masz na mysli ta sama ilosc permutacji w ogole jest ogladano podczas calego algorytmu i na trening zbiorze i na walidacyjnym?}

    \item \texttt{LocalE / GlobalE} \citep{lu2022fantastically}.
    Entropy-based probing for ordering selection. Given a candidate set of permutations, both methods rank permutations using the model's predictive entropy on a probing set and select the top-ranked ordering. \texttt{LocalE} uses the minimum entropy over the probing set (favoring permutations that yield at least one highly confident prediction), whereas \texttt{GlobalE} uses the average entropy (favoring permutations that are consistently confident).

    \item \texttt{PDO} \citep{xu2024context}.
    selects orderings using label-distribution criteria computed
    from the model’s predicted label probabilities. We report the three settings from the original work:
    \texttt{PDO-FewShot} (no additional data beyond the $k$ demonstrations),
    \texttt{PDO-FewShotU} (additionally uses an unlabeled set),
    and \texttt{PDO-FewShotUP} (uses an unlabeled set and a prior label distribution).

    \item \texttt{DEmO} \citep{guo2024makes}.
    performs a two-stage procedure: it first filters candidate orderings
    using a label-fairness criterion on content-free inputs, and then selects an ordering that maximizes
    an influence-based score for the test input (instance-level selection).
    
    \item \texttt{\ours{}} (ours: Section~\ref{sec:method}).
    We evaluate three variants of our approach:
    \begin{itemize}[leftmargin=*]
        \item \texttt{\ours{}-EMA:} a single PL model updated via an exponential moving-average (EMA) rank update.
        \item \texttt{\ours{}-1:} a single PL model fit by maximum-likelihood estimation (MLE) on the elite set.
        \item \texttt{\ours{}-4:} a mixture of $4$ PL components trained with an EM-style procedure.
    \end{itemize}
    We report the hyperparameters used for \texttt{\ours{}} in Appendix \ref{app:hyperparameters}.
\end{itemize}

%\swoboda{Skomentowac: Jak czesto sa odnoszenia do LLMow w LocalE/GlobalE/PDO,DEmO,PLO}.

\subsection{Results}

\paragraph{Classification.}
We evaluate on five classification benchmarks: MR (binary sentiment) \citep{pang2005seeing}, SST-5 (5-way sentiment) \citep{socher2013recursive}, TREC (question type) \citep{voorhees2000building}, AG's News (topic classification) \citep{zhang2015character}, and SUBJ (subjectivity) \citep{pang2004sentimental}. Table~\ref{tab:cls_results_k_qwen_llama_sibling} reports results for Qwen2.5-7B-Instruct \citep{yang2024qwen2} and Llama-3.1-8B-Instruct \citep{llama3modelcard}. Across datasets and number of ICL examples, \texttt{\ours{}} variants consistently rank among the top three methods on average.
We note that the difference to \texttt{Top-K} is small when few (e.g.\ 4) ICL examples are present, since then the number of permutations is small. We get larger performance differences once the space of permutations cannot be sampled effectively anymore, e.g.\ for $k \geq 8$.

\paragraph{Reasoning.}
We additionally evaluate \texttt{\ours{}} on the mathematical reasoning benchmarks—GSM8K~\citep{cobbe2021gsm8k}, MATH500~\citep{lightman2023lets}, and DeepMath~\citep{deepmath}.
Because these tasks have effectively unbounded answer spaces, label-probability based ordering methods such as \texttt{LocalE}/\texttt{GlobalE}, \texttt{PDO}, and \texttt{DEmO} are not directly applicable, so we compare against the Static and Top-K baselines. Table~\ref{tab:math_results_k_qwen} reports results for \texttt{Qwen2.5-7B-Instruct}. Across all three benchmarks and number of ICL examples \texttt{\ours{}} consistently outperforms \texttt{Top-K}, indicating that learning a distribution over demonstration orders remains effective beyond classification and transfers naturally to open-ended generation and reasoning settings.

\subsection{Ablations}

\begin{table}[ht]
\centering
\caption{Results for SUBJ across different number of $K$ of Mixture PL distribution and number of ICL examples $k$.}
\label{tab:placeholder_nk}
\renewcommand{\arraystretch}{1.00}
\resizebox{0.8\columnwidth}{!}{%
\begin{tabular}{c cccc}
\toprule
$K$ & $k=4$ & $k=8$ & $k=16$ & $k=32$ \\
\midrule
4  & 73.09 & 80.94 & 85.17 & 85.98 \\
8  &  73.04 & 80.10 & 84.17 & 86.75 \\
16 & 72.83 & 79.78 & 84.57 & 86.06 \\
32 & 72.52 & 80.48 & 83.95 & 86.25 \\
64 & 72.65 & 79.51 & 84.24 & 86.40 \\
\bottomrule
\end{tabular}%
}
\end{table}
\begin{table}[ht]
\centering
\caption{SUBJ results for \texttt{\ours{}} variants with weighted vs.\ unweighted MLE.}
\label{tab:subj_weighted_unweighted}
\renewcommand{\arraystretch}{1.2}
\setlength{\tabcolsep}{6pt}

\resizebox{\columnwidth}{!}{%
\begin{tabular}{l l cccc}
\toprule
Method & Weighting & $k{=}4$ & $k{=}8$ & $k{=}16$ & $k{=}32$ \\
\midrule
\multirow{2}{*}{\ours{}-1}
& Unweighted & 72.80 & 80.33 & 83.44 & 85.90 \\
& Weighted   & 72.78 & 80.88 & 83.85 & 87.48 \\
\midrule
\multirow{2}{*}{\ours{}-4}
& Unweighted & 73.09 & 80.94 & 85.17 & 85.98 \\
& Weighted   & 73.09 & 80.70 & 84.22 & 86.97 \\
\bottomrule
\end{tabular}%
}
\end{table}

\begin{figure}
    \centering
    \includegraphics[width=\columnwidth]{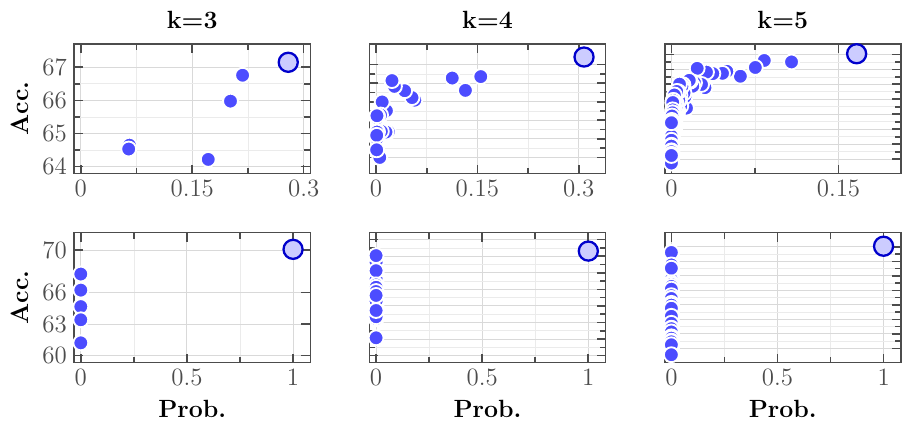}
    \caption{
    \underline{Top}: Ablation probability-test accuracy results for \texttt{\ours{}-EMA} and \underline{Bottom}: for \texttt{\ours{}-1}.
    }
    \label{fig:ablation}
\end{figure}

\begin{table*}[h]
\centering
\caption{Results for SUBJ for varying parameters $\rho$ of choosing the elite set in line~\ref{alg:elite-set-selection} in Algorithm~\ref{alg:pseudo-code}.}
\label{tab:elites}
\renewcommand{\arraystretch}{1.25}

\resizebox{0.9\textwidth}{!}{%
\begin{tabular}{c cccc cccc cccc}
\toprule
\multirow{2}{*}{$\rho$}
& \multicolumn{4}{c}{\ours{}-EMA}
& \multicolumn{4}{c}{\ours{}-1}
& \multicolumn{4}{c}{\ours{}-4} \\
\cmidrule(lr){2-5}\cmidrule(lr){6-9}\cmidrule(lr){10-13}
& $k=4$ & $k=8$ & $k=16$ & $k=32$
& $k=4$ & $k=8$ & $k=16$ & $k=32$
& $k=4$ & $k=8$ & $k=16$ & $k=32$ \\
\midrule
0.1 & 73.09    & 80.59    & 83.16    & 85.31    & 73.04  & 80.81  & 83.78  & 86.46  & 73.09  & 80.96  & 83.95  & 85.94  \\
0.2 & 72.78 & 80.64 & 83.98 & 86.75 & 72.80  & 80.33  & 83.44  & 85.90  & 73.09  & 80.94  & 85.17  & 85.98  \\
0.3 & 73.04    & 80.58    & 83.05    & 86.02    & 73.04  & 80.97 & 82.37 & 85.47 & 72.78  & 80.62 & 84.43 & 85.07 \\
0.5 & 72.75   & 79.91   & 82.65   & 84.90   & 72.91 & 79.83 & 82.06 & 86.15 & 72.91 & 79.81 & 82.00 & 83.88 \\
\bottomrule
\end{tabular}%
}
\end{table*}

\paragraph{Ablation: Number of PL components.}
We ablate the number of components $K$ in the mixture-of-Plackett--Luce model to quantify how additional mixture capacity affects downstream performance.
Intuitively, increasing $K$ should allow the learned ordering distribution to better capture multi-modal structure (i.e., multiple distinct high-performing permutations).
Table~\ref{tab:placeholder_nk} reports results on \textsc{subj}.
Overall, performance is largely stable across $K$, with only modest fluctuations.
This suggests that a small mixture (e.g., $K{=}4$) already captures the dominant modes of good orderings in this setting, and that further increasing $K$ yields diminishing returns.
We also see that potential overfitting occurs for larger $K$.

\paragraph{Ablation: Weighting in the MLE objective.}
We examine whether weighting elite permutations in the MLE refit improves performance. Concretely, we run \texttt{\ours{}} on \textsc{SUBJ} with both a single PL component and a $K{=}4$ mixture, comparing an \emph{unweighted} MLE objective (all elite permutations contribute equally) against a \emph{weighted} variant (elite permutations are reweighted by their training score). Results are shown in Table~\ref{tab:subj_weighted_unweighted}. Overall, the effect of weighting is inconsistent: depending on the number of demonstrations $k$ (and the PL capacity), weighting can yield small gains or slight regressions. Given the lack of a clear, consistent advantage, we adopt the unweighted MLE objective in all main experiments for simplicity.

\paragraph{Ablation: Proportion of elites.}
We study how the elite fraction $\rho$ (the proportion of sampled permutations retained for refitting in line~\ref{alg:elite-set-selection} in Algorithm~\ref{alg:pseudo-code}) affects performance.
We run all three \texttt{\ours{}} variants with varying $\rho$, and report results in Table~\ref{tab:elites}.
If $\rho$ is too large (e.g., $0.5$), the elite set becomes too broad and the update signal is diluted. If $\rho$ is too small (e.g., $0.1$), the update becomes overly selective and less stable. In both cases, the quality of the training signal deteriorates. Based on this trade-off, we use $\rho = 0.2$ in all experiments as a robust default.

\paragraph{Ablation: Relation between probabilities and test accuracy.}
A benefit of modeling orders with a PL distribution is interpretability: the learned probability of a permutation can be compared directly to its downstream utility. We test whether higher-probability permutations also achieve higher test accuracy. For small $k \in \{3,4,5\}$, we can enumerate all $k!$ permutations, compute their PL probabilities, and evaluate their true test accuracy. We run this analysis for \texttt{\ours{}-EMA} and \texttt{\ours{}-1}, averaging over 5 random seeds. In Figure~\ref{fig:ablation}, permutations are sorted from most to least probable and we plot the corresponding probability and test accuracy trends. Across all $k$, we observe a clear monotonic relationship: permutations assigned higher probability tend to yield higher test accuracy. We also find that \texttt{\ours{}-1} often concentrates mass on a single ordering (near-deterministic behavior), assigning probability close to $1$ to the best permutation and near $0$ to the rest, which is consistent with its MLE refitting objective and strong elite selection.

\section{Conclusions}
We have proposed an elegant probabilistic approach to choosing the ordering of ICL examples for prompting LLMs outperforming our reordering baselines.
We believe that our probabilistic perspective can be extended to obtain principled approaches also to other related problems, for example simultaneously selecting and ordering ICL examples from a large set, selecting an instruction prompt and reordering the ICL examples, etc.
Finally, reordering is largely absent from automatic prompt engineering approaches and we argue that this is another underexplored axis on how to improve automatic prompt engineering algorithms.

\newpage
\section{Limitations}
While \ours{} achieves strong results across diverse benchmarks, we note several limitations.

\begin{description}[leftmargin=!, labelwidth=10pt]
\item[Need for labeled data:]
\ours{} optimizes demonstration orderings using task-level metrics (e.g., accuracy), which requires labeled data to reliably score candidate permutations. This limits direct applicability in fully unsupervised settings.

\item[Task-specific optimization:]
In our current setup, \ours{} is optimized separately for each task. Developing a task-agnostic or transferable ordering strategy is an important direction for future work.

\item[Scaling to larger models:]
Our experiments focus on 7B--8B models. It remains to be seen whether the improvements persist at larger scales and under different inference regimes.
\end{description}

\section*{Ethics Statement}
We conducted this research in line with the ACL Code of Ethics and the ACM Code of Ethics and Professional Conduct.
Because our method optimizes the ordering of in-context demonstrations, it can amplify positional and recency effects: examples placed earlier may disproportionately steer model outputs, potentially reinforcing spurious correlations, biased behaviors, or unsafe instructions present in the prompt.
Order selection on a validation split can also introduce prompt-level overfitting and reduced robustness under distribution shift.
To mitigate these risks, we use disjoint splits for ordering selection and final evaluation, report results averaged over multiple random seeds and shot counts, and recommend filtering demonstrations for sensitive or unsafe content before applying ordering optimization.
More broadly, automated prompt optimization should be paired with robustness testing (including prompt-injection stress tests) and appropriate safeguards prior to deployment.

\appendix
\label{sec:appendix}

\section{Expressivity of Mixture-PL}
\begin{theorem}[Expressivity: Mixture-PL is dense, single PL is not (for $n\ge 3$)]
\label{thm:expressivity-mixture-PL}
Let $\Delta(S_n)$ denote the set of all probability distributions over permutations $S_n$ equipped with the $\ell_1$ norm
$\|p-q\|_1=\sum_{\pi\in S_n} |p(\pi)-q(\pi)|$.
\noindent\textbf{(i) Density of mixtures.}
For any target distribution $p\in \Delta(S_n)$ and any $\varepsilon>0$, there exist an integer $K\le n!$, weights
$\alpha_1,\dots,\alpha_K$ with $\alpha_k\ge 0$ and $\sum_{k=1}^K \alpha_k=1$, and logits
$\theta^{(1)},\dots,\theta^{(K)}\in\mathbbmss{R}^n$ such that the mixture
\[
q(\pi)\;=\;\sum_{k=1}^K \alpha_k\,\mathrm{PL}(\pi\mid \theta^{(k)})
\]
satisfies $\|q-p\|_1 < \varepsilon$. In particular, the family of mixture-PL models is dense in $\Delta(S_n)$.

\smallskip
\noindent\textbf{(ii) Non-density of a single PL component (when $n\ge 3$).}
For $n\ge 3$, the single-component PL family
\[
\mathcal{F}_{\mathrm{PL}}=\{\mathrm{PL}(\cdot\mid \theta):\theta\in\mathbb{R}^n\}
\]
is not dense in $\Delta(S_n)$.  
\end{theorem}

\section{Hyperparameters}
\label{app:hyperparameters}
In this section we give the exact hyperparameters we run our method with.
\smallskip

% =========================
% Hyperparameters (Appendix)
% =========================
\begin{table*}[ht]
\centering
\small
\setlength{\tabcolsep}{6pt}
\renewcommand{\arraystretch}{1.15}
\caption{Hyperparameters used for \ours{} (unless stated otherwise). We use the same notation as in the main body/Alg.~\ref{alg:pl_generic}: $T$ (CE iterations), $B$ (samples/iter), $\rho$ (elite fraction), $K$ (final draws), $\alpha$ (EMA smoothing), and $\tau$ (rank temperature). For the mixture model, $K$ also denotes the number of mixture components (as in the mixture definition in Sec.~\ref{sec:method}).}
\label{tab:hyperparameters}
\begin{tabular}{l c c p{7.2cm}}
\toprule
Hyperparameter & Symbol & Value & Meaning \\
\midrule
\multicolumn{4}{l}{\textbf{Experimental protocol}}\\
\# demonstrations & $k$ & $\{4,8,16,32\}$ & Shots used in the prompt. \\
\# seeds & --- & $5$ & Runs per dataset/$k$/method. \\
Validation budget & $|D_{\mathrm{val}}|$ & $1000$ & Total labeled pool used for ordering selection (if using a subset; otherwise \texttt{full}). \\
Inner/outer split & --- & $80/20$ & Inner set used for updates ($D_t^{\mathrm{train}}$); outer set used for final selection ($D^{\mathrm{val}}$). \\
Scoring batch size & --- & $16$ & Batch size for label-scoring / evaluation. \\
\midrule
\multicolumn{4}{l}{\texttt{\ours{} (shared CE / distributional optimization)}}\\
CE iterations & $T$ & $15$ & Number of distribution-update iterations. \\
Samples per iteration & $B$ & $15$ & Permutations sampled per iteration from the current distribution. \\
Elite fraction & $\rho$ & $0.2$ & Top-$\lceil \rho B\rceil$ permutations retained as elites. \\
Final draws & $K$ & $10$ & \# permutations sampled from the \emph{final} distribution for MC selection on $D^{\mathrm{val}}$. \\
EMA smoothing & $\alpha$ & $0.7$ & Smoothing used in the update (heuristic rank update and as optional smoothing for MLE/mixture refits). \\
Rank temperature & $\tau$ & $1.0$ & Rank-to-logit temperature in the heuristic update $\theta_i^{\mathrm{tgt}}=-\bar r_i/\tau$. \\
\midrule
\multicolumn{4}{l}{\texttt{\ours{}-1 (single PL, MLE refit on elites)}}\\
Adam steps & --- & $60$ & Gradient steps for PL MLE on the elite permutations. \\
Learning rate & --- & $0.1$ & Adam learning rate for PL MLE. \\
L2 penalty & --- & $0.0$ & L2 regularization coefficient in the MLE objective. \\
Logit clip & --- & $20$ & Clamp logits to $[-c,c]$ during MLE fitting for stability. \\
Weighted elites & --- & OFF & If ON, weight elite permutations by training score in the MLE fit. \\
\midrule
\multicolumn{4}{l}{\texttt{\ours{}-4 (mixture of PL, EM-style refit on elites)}}\\
\# mixture components & $K$ & $4$ & Number of PL components in the mixture. \\
Min component weight & --- & $10^{-3}$ & Lower bound on mixture weights during training (stability). \\
\bottomrule
\end{tabular}
\end{table*}

\section{Usage of LLMs}
We used LLMs to improve the clarity and style of the writing and presentation, and to assist with drafting parts of the implementation code. All substantive research decisions and the core technical contributions were made by the authors.

\end{document}